# Feynman on Artificial Intelligence and Machine Learning, with Updates

Eric Mjolsness

*Departments of Computer Science and Mathematics*
*University of California, Irvine*
emj@uci.edu

**Abstract**

I present my recollections of Richard Feynman's mid-1980s interest in artificial intelligence and neural networks, set in the technical context of the physics-related approaches to neural networks of that time. I attempt to evaluate his ideas in the light of the substantial advances in the field since then, and vice versa. There are aspects of Feynman's interests that I think have been largely achieved and others that remain excitingly open, notably in computational science, and potentially including the revival of symbolic methods therein.

## 1. Introduction

Richard Feynman's role in sparking the development of quantum computing is well known, as well as his affinity for calculating machines. Less well known is that he was seriously interested in artificial intelligence, and in neural networks in their 1980s incarnation as an approach to artificial intelligence, a subject led at Caltech by John Hopfield who had recently arrived as a full professor.

## 2. Selected Reminiscences

The Physics of Computation course began as a collaborative offering between Feynman, Hopfield, and Carver Mead, the pioneer of very large-scale integration (VLSI) computer chip design, who was also interested in neural computation as a new hardware paradigm. Feynman thought quite a bit about how to achieve progress in artificial intelligence (AI) that had eluded the mainstream symbolic approach, through neural networks instead, and particularly in their capabilities for pattern recognition and machine learning (ML).

As a physics graduate student at Caltech from 1980-1985, I was very fortunate to have many interactions with Richard Feynman, some of which bear on his interest in neural networks and artificial intelligence. To me these years at Caltech seemed like a time of great intellectual ferment in multiple overlapping directions. What follows is reconstructed from decades-old fallible memory. Others will almost surely remember things a bit differently.

Today neural networks are so successful that it is becoming hard to understand how non-mainstream such thinking was at the time. Neural networks had lost the evolutionary race with von Neumann computer architectures in the late 1950s and early 1960s, despite pioneering hardware work by Frank Rosenblatt, Bernard Widrow, and others. Enthusiasm and funding for neural network research had also suffered from the reception (but not the actual text) of the book on "Perceptrons" by Marvin Minsky and Seymour Papert which included negative results on what we could now call single-layer feed-forward neural networks without hidden units. But in the early 1980s John Hopfield, Geoffrey Hinton, Terry Sejnowski, Kunihiko Fukushima and others were trying to revive neural networks, with new quantitative dynamical algorithms and architectures. Feynman was interested (along with other physicists) because of the physics-like characteristics of this approach. The essence of a neural network could be described in quantitative terms, mostly by equations, at a higher level of abstraction than computer code. Feynman certainly knew Hopfield's work well; how much he knew about other neural network research is harder to say because he preferred to reinvent everything independently, from first principles.

At MIT Marvin Minsky was pioneer of artificial intelligence of the classic non-neural "symbolic" variety, and Feynman and Minsky were friends. As Feynman became interested in AI, somehow a deal was struck by which Minsky recommended that Papert's former doctoral student, Gerald Sussman, also prominent in mainstream symbolic AI as well as programming languages, should visit Caltech on an extended sabbatical where he could interact with Feynman on AI … and also, execute his plan to build a cost-effective special-purpose "digital orrery" planetary orbit computer. So Gerry Sussman collaborated on the Physics of Computation course too, delivering lectures on hardware, on programming using his own Scheme dialect of Lisp, and on AI.

The role in which I knew Feynman best was as a teaching assistant for the Physics of Computation class in several iterations from 1983-1985. Also, as a physics/neural network PhD student of John Hopfield, I was deputized to push the buttons and flip the tapes on a tape recorder for many of Feynman's lectures. Feynman was democratically-minded, and very often after class he would continue the discussion with his TAs and a changing cast of visitors including Sussman, at the Institutes' unpretentious cafeteria a few steps away. (US universities have evolved; the cafeteria became more pretentious.) Sometimes we would instead go to the Caltech Atheneum, the faculty club that also welcomed graduate students. Another TA for later iterations of this course was the remarkable physics graduate student Mike Douglas, then master of the digital orrery software and knowledgable in AI, later of string theory renown. Needless to say, the lunch discussions were fascinating and sometimes spectacular. The TAs sometimes also met Feynman in his office. I heard his voice ringing in my ears for years afterwards.

It is well known that Feynman had a leading role in organizing the computing effort at Los Alamos during the Manhattan Project in World War II, from 1943-1945. The "computers" of the day were human operators working on mechanical calculators, but the information flow among them had to be designed somehow. During the 1980s lunchtime socializing we heard previews of many of Feynman's Los Alamos and other stories that subsequently showed up in his later, memoir-like books. It is also well reported that in the mid-1980s, during the time of the Physics of Computation classes, he consulted for the Thinking Machines computer company. He worked on the bit-serial message-routing protocol for their highly parallel Connection Machine CM-1 computer and performed an analysis of its need for message overflow buffers. Also during this time, he undertook the quantum computing work described elsewhere in this book. These contemporaneous projects were described in the class. So his interest in computing was broad and long-lived.

Feynman had various dream projects that he didn't get time to finish. He died just a few years after giving these lectures, in 1988 at the age of 69. He wanted to think and visualize his way through the configuration space of quantum chromodynamics, the theory of the strong nuclear force, in order to deeply understand it the way he had visualized his way through several famous problems in statistical physics. And he wanted to train, shape, or perhaps evolve a neural network to achieve artificial intelligence, complete with vision and language. The neural network/learning machine (details were a bit sketchy) was apparently going to be person-like and female. "She" would master one skill after another, like a human baby, trained by doting scientists who would supply the right learning tasks in the right order.

Gerry Sussman, who was by far the most knowledgable AI expert on the AI-skeptical Caltech campus during his visit, did not love the Feynman plan for neural network style learning machines. He thought it was naive, harking back to the late-1950 – early-1960 ideas that had not panned out. On the other hand mainstream AI was just entering the "AI winter" phase of decreased confidence and funding, so Sussman could not win an argument on the future of AI outright, based on the accomplishments of the field at that time. Feynman in turn thought Sussman's "problem is he knows too much" about current AI research (he probably knew all of it, as the field was much smaller then), so Sussman couldn't see that the real progress was going to come from a completely different direction. These mutual diagnoses emerged only under fairly persistent questioning by the younger scientists present. In retrospect of more than 30 years, after the great strides in neural networks for computer vision and natural language of 2010 – 2020, Feynman looks to have had the decisive advantage in this symbolic vs. neural AI debate … but as I will suggest, things could change again!

### 3. Physics-like Neural Networks of the 1980s

We can use Feynman's approach to statistical mechanics to understand Hopfield's neural networks, and closely related networks. In particular, Feynman's variational method (e.g.[1], Section 3.4) allow us to understand key aspects of such neural networks. We will make a mathematical analogy between a network of idealized "neurons" that interact only with their neighbors in a network of connections, on the one hand, and a rigid lattice of simplified "atoms" that interact only with their neighbors in a solid material, on the other. The common theme is the use of real-number connection or interaction strength values $T_{ij}$ that specify whether, and how strongly, and with what $\pm$ sign two network nodes (either neurons or atoms, indexed by integers $i$ and $j$) influence one another. When $T_{ij} = 0$ then there is no network or lattice connection and no direct influence between nodes $i$ and $j$, and for large systems we can imagine this is the most common case. It is the nonzero connection strengths that specify the "network". A major point of difference between neural networks and solid material models in physics will be that the connection strengths in a neural network are allowed to change under some dynamics of learning.

For the physics-based material model we assume the Boltzmann probability distribution of equilibrium statistical mechanics: The probability of a state $p_{\text{Boltz}}(\text{state}) \propto e^{-\beta \text{Energy}(\text{state})}$. This proportionality makes sense because energy is extensive (additive over noninteracting subsystems), and probability is multiplicative over independent subsystems, such as networks that comprise unconnected subnetworks.

*3.1 Neural activation dynamics*

Next we establish the conventional statistical mechanics context, exhibit Feynman's convexity inequality, derive a mean field theory, and apply it to neural network activation dynamics.

For a Boltzmann distribution, the free energy $F(\beta, T, h) = \langle H \rangle - TS$ of an equilibrium statistical mechanics model on classical atomic "spin" variables $s_i \in \{\pm 1\}$ satisfies

$$e^{-\beta F} = Z(\beta, T, h) = \sum_{\{s_i \in \{\pm 1\}\}} e^{-\beta H[s]} , \qquad (1)$$

where for an Ising model, for example,

$$H_{\text{Ising}}[s] \equiv -\frac{1}{2}\sum_{i \neq j} T_{ij} s_i s_j - \sum_i h_i s_i . \qquad (2)$$

(In this notation $\beta = 1/T$ is the inverse temperature, angle brackets denote thermal averages with respect to the Boltzmann probability $p_{\text{Boltz}}[s] = e^{-\beta H[s]}/Z$, and $S = -\langle \log p[s] \rangle$ is entropy. We use units in which Boltzmann's constant $k_B$ is unity; otherwise we would have $\beta = 1/k_B T$. The connection strength matrix $\boldsymbol{T}$ is symmetric and may be sparse and/or structured. Don't confuse the indexed connection matrix $\boldsymbol{T} = [T_{ij}]$ with the scalar temperature parameter $T$.) To verify equation (2), it suffices to substitute $p_{\text{Boltz}}$ into $S$. We could also interpret the $s_i \in \{\pm 1\}$ variables as idealized binary-valued neuron number $i$ being "on" vs. "off", or "firing" vs. "not firing". Below we will seek energy-minimizing dynamics for "analog" real-valued neurons, whose real values are derived as probabilities of firing vs. not firing. This derivation proceeds by way of Mean Field Theory.

Since the free energy $F$ and the "partition function" $Z$ may be hard to calculate, we compare them to their values in some related system $(H_0, F_0)$ chosen to be easier to calculate:

$$Z = \frac{\sum_{\{s_i\}} e^{-\beta(H[s]-H_0[s])} e^{-\beta H_0[s]}}{\sum_{\{s_i\}} e^{-\beta H_0[s]}} e^{-\beta F_0} = \langle e^{-\beta(H-H_0)} \rangle_0 e^{-\beta F_0}$$

From the *convexity* of the exponential function, Feynman points out that

$$\langle e^{-\beta(H-H_0)} \rangle_0 \geq e^{-\beta \langle H-H_0 \rangle_0}$$

and therefore

$$e^{-\beta F} = Z = \langle e^{-\beta(H-H_0)} \rangle_0 e^{-\beta F_0} \geq e^{-\beta \langle H-H_0 \rangle_0} e^{-\beta F_0}$$

whence

$$F \leq F_0 + \langle H - H_0 \rangle_0 \quad \text{or} \quad F \leq \langle H \rangle_0 - TS_0 . \qquad (3)$$

Of course, Feynman actually does this for path integrals and quantum systems. This key inequality, relating quantities ($F$ and $Z$) that may be hard to calculate to quantities that are much easier to calculate, is sometimes attributed to Gibbs, Bogoliubov, and Feynman.

Now from equation (3) we can derive the Mean Field Theory (MFT) approximation choosing an independent product distribution of all the classical spins, i.e. $H_0 = \sum_i \mu_i s_i = -\sum_i u_i s_i$ . Then

$TS_0 = -\sum_i u_i \langle s_i \rangle_0 - F_0$ where $-\beta F_0 = \sum_i \log(2\cosh\beta u_i)$. The variational bound can be calculated as

$$F \leq \langle H \rangle_0 - TS_0 = -\frac{1}{2}\sum_{i \neq j} T_{ij}\langle s_i \rangle_0 \langle s_j \rangle_0 - \sum_i (h_i - u_i)\langle s_i \rangle_0 - \frac{1}{\beta}\sum_i \log(2\cosh\beta u_i)$$

Calculating $\langle s_i \rangle_0 = -\partial F_0 / \partial u_i$, or equivalently minimizing this upper bound with respect to $u_i$, yields $v_i \equiv \langle s_i \rangle_0 = \tanh(\beta u_i)$ where $v_i \in (-1,1)$. Then we can eliminate $u_i$ from variational bound in favor of $v_i$. The bound becomes:

$$F \leq E_{\text{MFT}}[v] \equiv -\frac{1}{2}\sum_{i \neq j} T_{ij} v_i v_j - \sum_i h_i v_i + \sum_i \varphi(v_i) \quad (4)$$

where $\varphi(v) = \frac{1}{\beta}[v\tanh^{-1}(v) - \log(2\cosh(\tanh^{-1}(v)))]$. Since $\tanh^{-1}(v) = \frac{1}{2}\log(\frac{1+v}{1-v})$ for $v \in (-1,1)$, we find

$$\varphi(v) = \frac{1}{\beta}[(\frac{1+v}{2})\log(\frac{1+v}{2}) + (\frac{1-v}{2})\log(\frac{1-v}{2})] = -TS_{\text{MFT}}[v],$$

which is another $\sum p\log p$ entropy expression since $(1 \pm v)/2 \in [0,1]$ and they sum to 1.

The quality of the approximation of equation (4) is naturally measured by how close the right hand side $E_{\text{MFT}}[v]$ comes to the left hand side quantity $F$, with the best possible approximation of this class being obtained by minimizing $E_{\text{MFT}}[v]$ as a function of all the $v_i$. We could do this algorithmically, or we could consider some *dynamics* for $v_i$ that seeks a minimum, and then implement that on a computer or more directly in analog electronic hardware.

So, in addition to fast equilibration of the vector $\mathbf{s} = [s]$ of spins $s_i$, we now imagine a slower time-scale dissipative dynamics that acts irreversibly to minimize the approximating free energy as modeled by $E_{\text{MFT}}[v]$. The derivative of the potential $\varphi(v)$ is $\tanh^{-1}(v)/\beta$, so $E_{\text{MFT}}[v]$ is minimized at $v_i = \tanh(\beta u_i \equiv -\beta \partial E_{Ising}[v]/\partial v_i)$, i.e.

$$v_i = \tanh(\beta \sum_j T_{ij} v_j + \beta h_i) \quad (5)$$

at a local minimum. Once an energy-minimizing system has reached an isolated local minimum of the energy, it can go no further and must stop - it has reached a "fixed point" of the dynamics. So equation (5) is the fixed-point equation for Hopfield's "analog" (real-valued, $v_i \in (-1,1)$) neural network. Similarly, for MFT and $\{0,1\}$-valued "neurons" or "units" we obtain the entropy expression $\varphi(v) = \frac{1}{\beta}(v\log v + (1-v)\log(1-v))$, whence $v = g(u) = e^{\beta u}/(1+e^{\beta u})) = 1/(1+e^{-\beta u})$.

More generally, $\varphi(v) = \int^v g^{-1}(v)dv$ where $g$ is the artificial neuron's (monotonic) activation function. Then the fixed-point equation becomes

$$v_i = g_i(\sum_j T_{ij} v_j + h_i) . \quad (6)$$

One can also allow the potential function $\varphi$ and the activation function $g = \varphi'^{-1}$ to depend on the variable index $i$ as indicated in equation (6), while retaining the useful energy-minimizing optimization property of the network with $E[v] = -\frac{1}{2}\sum_{i \neq j} T_{ij} v_i v_j - \sum_i h_i v_i + \sum_i \varphi_i(v_i)$, *but either of these generalizations gives up the derivation from equilibrium MFT.*

Hopfield also introduced a descent dynamics capable of finding these fixed-points by energy minimization [2]:

$$v_i = g(u_i) \quad \text{and} \quad \tau_i \frac{du_i}{dt} + u_i = \sum_j T_{ij} v_j + h_i . \tag{7}$$

Input to the system can be provided through $\mathbf{h}$ or through the initial condition $\mathbf{v}(t = 0)$. This architecture was applied to small combinatorial optimization problems, such as the Traveling Salesman Problem.

This variational MFT derivation generalizes naturally [3] from Ising "spins" $s_i \in \{\pm 1\}$ to Potts model discrete state variables which take one of a fixed finite number of values $q_i \in \{1, \ldots A\}$ that we can encode using $s_{ia} \in \{0,1\}$ and $q_i = \sum_a a s_{ia}$, where $\sum_a s_{ia} = 1$. Then the potential energy function becomes $\varphi([v_{i*}]) = \frac{1}{\beta} \sum_a v_{ia} \log v_{ia} = -T\tilde{S}[v]$ (the entropy of the $i^{th}$ soft choice distribution) and the activation function is:

$$v_{ia} = e^{\beta u_{ia}} / \sum_b e^{\beta u_{ib}} , \quad \text{where} \quad u_{ia} \equiv \partial E_{Potts}[v] / \partial v_{ia} . \tag{8}$$

This is the well-known "soft-max" activation function, which softly (depending on parameter $\beta$) chooses among a finite set of alternatives so as to maximize $u_{ia}$ over index $a$, automatically obeying the constraints $v_{ia} \geq 0$ and $\sum_a v_{ia} = 1$. Its usage and importance have steadily increased over time since it provides a differentiable choice over more than two alternatives. A further natural generalization is to the "soft-assign" fast dynamics which alternates row and column normalizations like equation (8) and converges to a solution satisfying the constraints of a doubly stochastic or "assignment" matrix: $v_{ia} \geq 0, \sum_a v_{ia} = 1$, and $\sum_i v_{ia} = 1$. The soft-assign architecture was applied to a number of semi-symbolic problems (e.g. [4]) bordering on the cognitive.

*3.2 Neural learning dynamics*

In a somewhat looser manner, we can use statistical mechanics ideas to obtain *learning dynamics* in neural networks. We allow the symmetric, real-valued connection matrix $T$ to become dynamic by analogy with the activation spin variables, adding a potential function $\Phi(T_{ij})$ chosen here so the resulting activation function $\Phi'^{-1}$ for $T_{ij}$ will be linear:

$$H[s,T] = -A \sum_i h_i s_i - B \sum_{i<j} T_{ij} s_i s_j + B \sum_{i<j} \Phi(T_{ij}) \tag{9}$$

$$\Phi(T) = \frac{1}{2c} T^2$$

where the parameters $A \geq 0$ and $B \geq 0$ have different relative sizes in learning mode vs. recall mode. In learning mode $A \gg B$ is large so $\mathbf{s}$ tracks and binarizes the externally imposed input vector $\mathbf{h}$, and $T$ adapts to that input. At low temperature, energy is minimized and the matrix entry $T_{ij}$ is driven down the gradient $\partial H / \partial T_{ij}$ towards $T_{ij} = c s_i s_j$. For example, one may use dissipative dynamics

$$\tau_T \frac{dT_{ij}}{dt} + T_{ij} = c s_i s_j . \tag{10}$$

If members of a small set of input vectors $\{h\}$ are imposed repetitively, and the learning rate $1/\tau_T$ is low enough, then the binarized input vectors $s$ will get averaged over the input population: $T_{ij} \to c\langle s_i s_j\rangle_{\text{population}}$. On the other hand, in recall mode, $B \gg A$ and the learning rate is zero, so $\dot T = 0$ and at low temperature the activation state $s$ falls into whichever local minimum of $-\sum_{i<j} T_{ij} s_i s_j$ best aligns with the current input $h$ or $s(t=0)$.

The resulting computational functionality is that of a "content addressable memory", which retrieves a learned memory by its similarity to an input vector, as was demonstrated computationally for a discrete-time version of this learning rule in [5]. It has similarities to the Boltzmann machine learning algorithm [6], and even to Rosenblatt's (1962) perceptron learning rule. Equations like (10) that model connections or "synapses" $T_{ij}$ as strengthening or weakening depending on coincidence or correlation between the "neurons" they connect, also realize "Hebb's rule" for synaptic plasticity mediating learning as proposed in neurobiology.

Another important learning rule can be derived as follows, starting with equations (6) and/or (7). If we impose a sparsity pattern on the weights $T$ by partitioning the neurons (units) into *layers* indexed by $l \in 0, \dots L$, with nonzero connection weights only allowed between adjacent layers $(l-1, l)$ (or more generally from earlier to later layers), and if we further scale such weights by a factor of $\varepsilon^l$ where $\varepsilon \to 0$, and allow a compensating layer-dependent $\beta \propto \varepsilon^l$ parameter in the activation functions $g_l(v_{il})$, *then* layer $l$ effectively only receives input from layer $l-1$ and not from layer $l+1$. The result is a feed-forward multilayer neural network, often called a "multilayer perceptron" (MLP), with solution dynamics obtained simply by updating the fixed-point equation

$$v_{il} = g(u_{il}) \text{ where } u_{il} = \sum_{j=1}^{J_l} T_{ij}^{(l)} v_{jl-1} + h_{jl-1} = \sum_{j=1}^{J_l+1} T_{ij}^{(l)} v_{jl-1}, \qquad (11)$$

once for each layer in succession, ordered by increasing layer number. Here the convention $v_{J_l+1\, l} \equiv 1$ and $T^{(l)}_{iJ_l+1} \equiv h_{jl-1}$ obviates the need for a separate learning rule for the biases $h_{jl}$.

Now learning can proceed by following a gradient with respect to the weights $T_{ij}$, that can be efficiently calculated. One imposes an objective function on the output activations $v_{jL}$ in the last layer, such as $E(T) = \sum_p E_p(T)$ where $E_p(T) = \frac{1}{2}\sum_i (v_{iL} - y_{pi})^2$, and where $p$ indexes input pattern vectors $x_i = v_{i0}^{(p)}$ and we have suppressed extra $(p)$ indices on the activations. Now we just need to use the chain rule to find all the derivatives recursively, by assuming every $u$ and $v$ activation variable depends on all the weights of its own and earlier layers. By this means we will learn an approximate mapping (function) from input vector $x$ to output vector $y$.

For any layer $l > 0$ we can differentiate $u_{il}$ in equation (11) to find
$$\frac{\partial E_p}{\partial T_{ij}^{(l)}} = \frac{\partial E_p}{\partial u_{il}} \frac{\partial u_{il}}{\partial T_{ij}^{(l)}} \equiv \delta_i^{(l)} v_{jl-1}$$
(again suppressing $p$ indices). We need to find the $\delta$s. For output layer $L$ in particular, we can differentiate $E_p$ and $v_{il}$ in equation (11) to find

$$\delta_i^{(L)} \equiv \frac{\partial E_p}{\partial u_{iL}} = \frac{\partial E_p}{\partial v_{iL}} \frac{\partial v_{iL}}{\partial u_{iL}} = (v_{iL} - y_{pi})g'(u_{iL}) \ . \tag{12}$$

For any other layer $l - 1 \geq 0$ except the last one, we can differentiate equation (11) to find

$$\delta_i^{(l-1)} \equiv \frac{\partial E_p}{\partial u_{il-1}} = \sum_k \frac{\partial E_p}{\partial u_{kl}} \frac{\partial u_{kl}}{\partial u_{il-1}} = g'(u_{il-1})\sum_k \delta_k^{(l)} T_{ki}^{(l)}$$

i.e.

$$\delta_i^{(l-1)} = g'(u_{il-1})\sum_k T_{ik}^{(l)\text{transpose}} \delta_k^{(l)} \ , \tag{13}$$

in which the sum over $k$ is a matrix multiplication in the $(i, k)$ index space. This method is sometimes called the "generalized delta rule". It emerges from this calculation that the crucial error signal $\Delta_{il}$ travels *backwards* through the layers of the network, by the transpose of each layer's weight matrix. Hence the algorithm's name, "backwards propagation of error" or just "backpropagation".

Now we have the ingredients for continuous descent dynamics $\tau_T dT_{ij}^{(l)}/dt = -\eta_{\text{cont}} \partial E / \partial T_{ij}^{(l)}$
$= -\eta_{\text{cont}} \Sigma_p \partial E_p / \partial T_{ij}^{(l)} = -\eta_{\text{cont}} \Sigma_p \delta_i^{(pl)} v_{jl-1}^{(p)}$, or discretized in time to first order:

$$\Delta T_{ij}^{(l)} = \Sigma_p \Delta_p T_{ij}^{(l)} = -\eta \Sigma_p \delta_i^{(pl)} v_{jl-1}^{(p)} \tag{14}$$

where $\eta = \eta_{\text{cont}} \Delta t / \tau_T > 0$. The more cost-effective "stochastic gradient descent" version of this rule is just to update according to one input pattern $p$ at a time as in equation (10): $\Delta T_{ij}^{(l)} = \Delta_p T_{ij}^{(l)} = -\eta \Delta_{il}^{(p)} v_{jl-1}^{(p)}$.

Equation (13) is pregnant with consequences. It implies that a many-layered ("deep") network will have both products of many factors of $g'$, each of which can be quite large or small, and many factors of $T^{(l)\text{transpose}}$ potentially resulting in a large matrix condition number. The resulting numerical problems fomented an accumulation special methods or "hacks" to deal with them. One of the most common is weight decay: adding a regularizer $\frac{\lambda}{2}\Sigma_{lij}(T_{ij}^{(l)})^2$ to the objective $E$, whose negative derivative then shows up in the gradient descent dynamics for learning.

Besides the layer-stratified weight matrix $T(\varepsilon \to 0)$ mapping used above, there is another, more rigid way to map Hopfield analog neural networks into multilayer feed-forward neural networks: one can "unroll" or "unfold" them by adding a discretized time index. For example, a forward Euler solution method for equation (7) could be expressed as

$$v_{ik} = g(u_{ik}) \quad \text{and} \quad \tau_i \frac{u_{ik} - u_{ik-1}}{\Delta t} + u_{ik-1} = \sum_j T_{ij} v_{jk-1} + h_i + O(\Delta t^2)$$

and approximated by discrete-time dynamics

$$v_{ik} = g(u_{ik}) \quad \text{and} \quad u_{ik} = (1 - \frac{\Delta t}{\tau_i})u_{ik-1} + \frac{\Delta t}{\tau_i}(\sum_j T_{ij} v_{jk-1} + h_i) \tag{15}$$

Even if $\tau_i \neq \Delta t$ equation (15) is very close to the form of equation (11), especially after absorbing the relative time factors into $T$ and $h$. So this architecture supports a modified "delta rule" for learning by backpropagation of error. Alternatively, there are several slightly generalized forms of (10) that can provide a common generalization of (11) and (15), complete with a delta rule for backpropagation learning. For example, one can add extra linear units ($g_i(u_i) = u_i$) to convey the $\boldsymbol{u}$ information of layer $l-1$ forward to layer $l$. The more substantial difference between (15) and (11) is actually one of specialization: In (15), all layers have the same connection matrix $T_{ij}^{(l)} = T_{ij}$, whereas in (11) they generally differ.

This reduction of parameters is one example of "weight sharing", whereby the full connection matrix (respecting an architecture's pattern of sparsity) is a function of some smaller number of parameters $\theta$, and the chain rule is used to create a version of backpropagation that computes gradients with respect to $\theta$ from backpropagated gradients with respect to free weights $T$.

In this way we obtain a family of gradient descent learning algorithms for Hopfield-style continuous time analog neural networks with feedback, a type of "recurrent neural network". Better ODE integration schemes will give rise to somewhat more complex learning algorithms. The method can be generalized to other differential equations with other kinds of parameters than neural networks. In the continuous limit $\Delta t \to 0$ we recover the "adjoint method" for locally optimizing parameters in a differential equation model, again with the training signals flowing backward in time.

## 4. The AI/ML Spring

So, what has changed since the 1980s? And which changes would have resonated with Feynman?

Neural networks are usually specified with a combination of algebraic equations with free (learnable) parameters, and illustrated by block diagrams showing how variables are connected to one another in these equations. The total amount of information to specify the architecture is usually very small indeed, compared to that required by a computer program aimed at a similarly nontrivial problem in perception, pattern recognition, or many other successful applications. Almost all of the information in a trained network comes instead from the data set used to train it, and is stored in the large number of learned parameters: the numerical "weights" by which one artificial neuron or "unit" influences another.

The field of neural networks has gone through several transformations since the 1980s. Some of the innovations that can be expressed in simple equations include:
- Fukushima's neurobiologically inspired "neocognitron" networks,
- Hopfield's optimizing or energy-minimizing networks (e.g. equation (7) above),
- Boltzmann (probabilistic) machines,
- Backpropagation of error through feed-forward multilayer perceptrons (variants and specializations of equations (11) and (13) above),
- Deep convolutional networks (essentially, a neocognitron network with backpropagation of error learning),
- "Soft-max" mixtures of experts,

- "Soft-assign" correspondence-optimizing networks,
- Regularization methods such as "weight decay" and random unit "dropout" that promote successful generalization,
- Restricted Boltzmann machines,
- Graphical probabilistic models,
- Latent semantic spaces,
- Variational autoencoders,
- Long short term memory,
- Residual networks,
- Tensor networks,
- "Transformer" architectures based on soft-max "attention" equations incorporating equation (8) above,
- Graph neural networks, and
- Multiscale/multigrid neural networks,

among many others.

Other kinds of learning architectures that are mathematical and specified by equations, without necessarily being neural networks, include:
- Self-organizing maps,
- Kernel methods,
- Manifold learning, and
- Information geometry.

Often practitioners encounter these architectures in the form of computer code or somewhat ill-defined diagrams. But what is essential to specify are the relatively compact equations, and ample training data. This combination sounds like theoretical and experimental physics. In an odd temporal coincidence, and as shown in this book, both Feynman and Hopfield wrote down Hamiltonians as simple equations describing nontraditional computing machines; Feynman for quantum computing (1986, and presented earlier in the Physics of Computation class), and Hopfield for neural networks (1982). The Boltzmann Machine of Ackley, Hinton and Sejnowski (1985), too, has a statistical physics Hamiltonian description.

Other aspects of neural network performance enhancement are more procedural than mathematical. One procedural approach to regularization is "early stopping" of a training optimization algorithm. Another example is Feynman's advocacy of a carefully designed sequence of learning tasks, related to what is known in animal training as "shaping". This occurs in current neural network practice as "transfer learning" and is very effective across many domains - although the sequence may be less important than the collection of learning tasks on which the same network is jointly trained ("multi-task learning") - and also to "lifelong learning" which is more or less what it sounds like. And paradigms such as reinforcement learning and unsupervised learning each start with a different premise about how data from the world is to enter into the training process.

At this point I must apologize to my many colleagues whose own great architectures and algorithms did not happen to get onto the foregoing lists. I am barging through decades of high-quality research, and laying waste to nuance.

With these machine learning architectures and procedures, vast improvements have taken place in the decade 2010-2020. As is now widely known, these improvements have resculpted the classic AI areas of computer vision, natural language, robotics, game-playing, and recently even theorem-proving which was the *ne plus ultra* of symbolic AI (Newell and Simon's Logic Theorist, 1956; Robinson's resolution-based theorem proving, 1965). Commercial and industrial applications are legion. There are now many domains in which it is not possible to be competitive without using machine learning. Even the practice of the hard sciences, including physics, is starting to feel an influx of AI/ML methods adapted to include scientific knowledge, under banners such as "physics-informed machine learning".

Some of the best known and most influential recent successes in neural networks include AlexNet [7] which was a milestone in computer vision, AlphaGo [8] and AlphaZero (2017) for playing the intellectually demanding board games of Go and Chess, and the Generative Pre-Trained Transformer GPT-2 [9] that generates readable natural language text in response to natural language prompts. These examples go far towards answering classic challenges that were identified right from the beginning of AI (for example, early computer vision at MIT, "Los Alamos chess" without bishops on a 6x6 board, and the "SHRDLU" and "ELIZA" text interaction programs - all running on computers of miniscule capacity by current standards). These notable examples involve neural network training of tens of millions to tens of billions of parameters, and they comprise a complete revolution in the field by comparison to AI capabilities in the mid-1980s. But more importantly, using the kinds of techniques listed above, there is now a vast array of successful applications. An increasing fraction of high-technology industry, including the hardware and software that we all interact with every day, is dependent on neural networks and related machine learning trainable models. Consequently, a very large amount of computing goes into their training every day.

What would Feynman have thought of all all these advances? Remembering his commitment to having a personal, independent point of view on everything, and the impossibility of emulating a great mind, there can be little certainty on this point. Nevertheless it seems to me that:

1. He would be very comfortable with the data-driven nature of the field, which was developing but far from universal in pre-1980s symbolic AI, and with the fact that serious validating computer experiments with some kind of "real data" are universally expected as part of the development of new techniques. This characteristic may be partly due to the large influx of physicists and other quantitatively-minded people into the field.
2. However, Feynman reviewed experiments critically. He expounded that in science "you are the easiest one [for you] to fool", in his classes and public lectures. I think he might be guarded about AI/ML methodological and sociological problems such as stopping upon success, selective reporting of best-among-competitors performance, community overtraining on benchmarks, and related sources of what he called "too much happy talk" in science.

3. He would be enthusiastic about transfer learning, and about the particular neural architectures that have led to success in vision, natural language, and limited combinations thereof. These methods have essentially achieved the goals of the machine learning project that he had in mind - insofar as he had expressed them.
4. Regarding equations as the hidden currency of learning architecture, he might be of two minds: attracted because he was a mathematical master; repelled because the source of his mastery was the ability to deeply visualize the meaning of each equation. Neural network equations for the most part are just not that conceptually deep. In the old AI dichotomy of "scruffy" vs. "neat" research approaches, ML equations might be neat but not neat enough.
5. He would not be too worried about the large number of small, apparently unconnected hacks required to get all these successes, nor about their lack of connection to the neurobiology of living "neural" networks. For example, it was just an empirical matter to him, whether computer chess programs would eventually work better by brute force search or by expert-derived heuristics. In the old AI dichotomy of "scruffy" vs. "neat" approaches, "scruffy" but functional is fine, as advocated by Minsky.
6. He might nevertheless be intrigued by conceptually deeper, less *ad hoc* architectures that connect to physics, such as manifold learning (clearly related to general relativity), and perhaps by attempts to connect neural networks to real neurobiology.
7. He might be skeptical of the "causal inference by structural equations" approach to infusing AI/ML into reasoning in the softer sciences, because it has no real dynamics and it does not address the main causality problem, namely of fitting into the multiscale and reductionist perspective on science that he expounded as early as his 1964 Messenger lectures [10]. At a fundamental level at least, "causality" is better stated in quantum field theories by Feynman propagators and by the spacelike commutation and anticommutation of operators.
8. Because of his deep commitment to physics, Feynman would be quite interested in - if also a bit skeptical of - the "physics-informed machine learning" agenda which is now being pursued with many different architectures, representations, and methods for the purpose of doing computational physics, chemistry, and biology. If I had to pick one thing about present-day neural networks, machine learning, and AI that Richard Feynman would be most interested in, this would be it.
9. He would *certainly* have creative and potentially powerful new ideas that are not yet on anybody's list or agenda.

**5. AI/ML for Computational Science**

To a theoretically-minded scientist, one of the most exciting things happening in AI/ML today (as I write in 2021) is its burgeoning adoption into computational science. The interesting neural network architectures for this purpose are those that incorporate physical or other scientific principles and constraints into their searchable function spaces. For example, many-electron quantum mechanics can be tackled using variational methods (of which Feynman was a master, e.g. [1]) with wave function bases that incorporate Slater determinants for permutation antisymmetry; one neural network generalization of this method exists under the name of "FermiNet" and has shown promise [11]. Another ML-enriched way to compute quantum potential energy functions is by training neural networks to approximate them on many density functional theory quantum electron ground state calculations [12].

Working our way up the scale hierarchy that Feynman liked to describe, a standard method for simulating multi-molecular systems in biology, biochemistry, and materials science is using Molecular Dynamics (MD) particle-based (non-quantum) stochastic dynamics. Methods for speeding up such simulations using machine learning now exist [13]. What is more, methods like ANI-1 or FermiNet neural networks are capable of learning potential energy functions that could appear in MD simulations, creating a truly multiscale model stack, which Feynman thought was important as "this tremendous world of interconnecting hierarchies" but worried that the inter-level connections were actually "… a little weak. We have not thought them all through yet." [10]. But now perhaps we can think them through, aided by artificial intelligence.

A considerably further jump up the scale hierarchy leads theorists to spatial continuum models modeled as partial differential equations, for example for fluid flow or elastic mechanics. The solution of these and many other PDEs can now be vastly accelerated, and indeed the PDEs can themselves be learned from data, using a variety of different neural network and related machine learning methods [14]. Related work in multiscale model reduction using ML, about which I am reasonably hopeful, includes my own collaborative efforts that use a dynamical version of the Boltzmann Machine learning algorithm on additive fundamental process operators [15], and/or graph neural networks, for that purpose. Other examples of ML for computational science are included in the review [16].

Of course, science is grounded in experiment and observation that produce data, all aspects of which are also being revolutionized by machine learning [17], and codified in deep theoretical structures which may yet be impacted as well [18]. But I think the current flowering of AI/ML for multiscale, "hierarchical" computational science would have had a strong appeal and resonance for Richard Feynman and might have attracted his unique efforts.

## 6. A Mathematical Synthesis, and Return to Symbolic AI?

Could the pendulum ever swing back from numeric AI towards symbolic AI? One motivation for that possibility is the widely cited need for "explainable" versions of ML models. Another is to remember that the actual mathematical specification of most ML methods is most clearly and concisely done by (symbolic) equations. Such ML equations can be derived from statistical models and automatically differentiated and parameter-optimized, although it is now possible to automatically differentiate and optimize some generic computer programs as well. So the pendulum could swing back, at least if it takes a mathematical trajectory and fully absorbs current ML methods. In that connection, there are presently two main strands of symbolic computing centered on mathematics.

Minsky's former doctoral student Joel Moses led the 1970's charge at MIT into computer algebra in the form of the Macsyma program which was capable in integral calculus, power series manipulation, solution of differential equations, and many other useful applied mathematical methods. Computer algebra (CA) systems propagated modestly and comprised a continuously successful niche for symbolic computing with an "AI" flavor; they did not suffer an AI winter. At a slower burn that achieved academic but not commercial success, automatic theorem proving (ATP) systems kept improving over the intervening decades as well - for example there was an entire Journal of Formalized Mathematics, which automatically verified all submitted papers. More recently a number of substantial results in mathematics such as the Kepler sphere-packing conjecture [19] have been proved using a new generation of Interactive Theorem Verification (ITV) systems with powerful inference engines such as "hammers" that use automated theorem proving to suggest verifiable proof steps in ITV, and scalable "satisfiability modulo theories" logical solver algorithms. So in the realm of mathematics, symbolic AI is alive and well on a very modest scale. Could symbolic AI break out from this small but critical niche (CA and/or ITV for automating mathematics), into scientific computing (which tends to be mathematical) or into general computing?

The only viable strategy for competing with ML methods at this point is to co-opt them. Fortunately, the way is clear to combining ML with both CA and ITV as shown by some recent papers that may soon be superseded by even stronger results. In the realm of computer algebra for symbolic integration and differential-equation solving, there is good progress in approaching these problems by means of deep neural networks [20], although more remains to be done [21]. ML has been applied to improving the ATP inference engines in ITV [22, 23, 24, 25, 26] based on deep reinforcement learning, GPT-2, graph neural networks, and $k$-nearest neighbors respectively, as well as on co-training on multiple related tasks. Conversely, ITV can be used to verify ML theory (e.g. [27]) and potentially to improve the reliable generation of ML code.

Of course, better proof alone is not enough. Feynman was a mathematical innovator, but in several lectures he expressed the physicists' sentiment that "It's possible to *know* more than you can *prove*". Another symbolic method is "symbolic regression", including an interesting experiment in rediscovering physics formulae appearing in Feynman's milestone lectures on physics [18]. My own as-yet informal proposal for deep symbolic/ML integration in computational science is outlined at the end of [28]. Together with ML in CA and ITV, these various recent works point to the possibility of powerful synergistic combinations of symbolic and numerical approaches to mathematical AI, perhaps first in applied-mathematical domains such as computational science, but eventually in general computing as well.

## 7. Conclusion

In summary, many of Richard Feynman's ideas about the future of neural network-like machine learning systems have come to pass. Since these systems work well now, he would probably have devoted time and energy to finding the next deep conceptual breakthrough they enable - particularly in the sciences.

## References

[1] Feynman, R.P. *Statistical Mechanics*; Reading MA: Benjamin, TX, USA, 1972. https://hdl.handle.net/2027/uc1.b4227910